\documentclass[10pt,twocolumn,letterpaper]{article}

\usepackage{cvpr}
\usepackage{times}
\usepackage{epsfig}
\usepackage{graphicx}
\usepackage{amsmath,graphicx}
\usepackage{amssymb}
\usepackage[numbers,sort&compress,sectionbib]{natbib}
\usepackage{amsfonts}
\usepackage{acro}
\usepackage{mathtools}
\usepackage{mathrsfs}
\usepackage{multirow}
\usepackage{authblk}

\newcommand{\Sec}{Sec.}
\newcommand{\Fig}{Figure}
\newcommand{\Eq}{Eq.}

\newcommand{\Tab}{Table}

\DeclareAcronym{oar}{
short=OAR,
long=organ at risk,
long-plural-form=organs at risk,
}

\DeclareAcronym{hn}{
short=H\&N,
long=head and neck
}

\DeclareAcronym{sh}{
short=S\&H,
long=small \& hard
}

\DeclareAcronym{CT}{
short=CT,
long=computed tomography
}

\DeclareAcronym{phnn}{
short=P-HNN,
long=progressive holistic-nested network
}

\DeclareAcronym{RTCT}{
short=RTCT,
long=radiotherapy computed tomography
}

\DeclareAcronym{p3d}{
short=P3D,
long=Pseudo-3D
}

\DeclareAcronym{nas}{
short=NAS,
long=neural architecture search
}

\DeclareAcronym{dsc}{
short=DSC,
long=Dice score
}

\DeclareAcronym{hd}{
short=HD,
long=Hausdorff distance
}

\DeclareAcronym{hd95}{
short=HD95,
long=Hausdorff distance 95\%
}

\DeclareAcronym{asd}{
short=ASD,
long=average surface distance
}

\DeclareAcronym{cnn}{
short=CNN,
long=convolutional neural network,
}

\DeclareAcronym{fcn}{
short=FCN,
long=fully convolutional network,
}

\DeclareAcronym{voi}{
short=VOI,
long=volume of interest,
long-plural-form=volumes of interest,
}

\DeclareAcronym{soars}{
short=SOARS,
long=stratified organ at risk segmentation,
}

\DeclareAcronym{sota}{
short=SOTA,
long=state-of-the-art,
}

\newcommand{\guo}[1]{{\color{black}#1}}

\usepackage[pagebackref=true,breaklinks=true,letterpaper=true,colorlinks,bookmarks=false]{hyperref}

\cvprfinalcopy % *** Uncomment this line for the final submission

\def\cvprPaperID{2384} % *** Enter the CVPR Paper ID here

\ifcvprfinal\fi

\begin{document}

%%%%%%%%% TITLE

\title{Organ at Risk Segmentation for Head and Neck Cancer using Stratified Learning and Neural Architecture Search}

\author{Dazhou Guo$^{1}$ \,\, Dakai Jin$^{1}$ \,\, Zhuotun Zhu$^{3}$ \,\, Tsung-Ying Ho$^{2}$ \,\, Adam P. Harrison$^{1}$ \,\, Chun-Hung Chao$^{4}$ \,\, Jing Xiao$^{5}$ \,\, Alan Yuille$^{3}$ \,\, Chien-Yu Lin$^{2}$ \,\, Le Lu$^{1}$\\

\small{$^1$PAII Inc. \,\, $^2$Chang Gung Memorial Hospital \,\, $^3$The Johns Hopkins University \,\, $^4$National Tsing Hua University \,\, $^5$Ping An Technology}
\thanks{\textit{\textbf{This paper is accepted by CVPR 2020. The final published version of the proceedings is available on IEEE Xplore.}}}
}

\maketitle
%\thispagestyle{empty}

%%%%%%%%% ABSTRACThttps://www.overleaf.com/project/5dadf1685b59f20001661bba

\begin{abstract}
\vspace{-1em}
   \Ac{oar} segmentation is a critical step in radiotherapy of \ac{hn} cancer, where 
   inconsistencies across radiation oncologists and prohibitive labor costs motivate automated approaches. However, leading methods using standard \acl{fcn} workflows that are challenged when the number of \acp{oar} becomes large, \eg $>$ 40. For such scenarios, insights can be gained from the stratification approaches seen in manual clinical \ac{oar} delineation. This is the goal of our work, where we introduce \ac{soars}, an approach that stratifies \acp{oar} into anchor, mid-level, and \ac{sh} categories. \ac{soars} stratifies across two dimensions. The first dimension is that distinct processing frameworks are used for each \ac{oar} category. In particular, inspired by clinical practices, anchor \acp{oar} are used to guide the mid-level and \ac{sh} categories. The second dimension is that distinct network architectures are used to manage the significant contrast, size, and anatomy variations between different \acp{oar}.  We use differentiable \ac{nas}, allowing the network to choose among 2D, 3D or Pseudo-3D convolutions. Extensive 4-fold cross-validation on $142$ \ac{hn} cancer patients with $42$ manually labeled \acp{oar}, \emph{the most comprehensive \ac{oar} dataset to date}, demonstrates that both framework- and  \ac{nas}-stratification significantly improves quantitative performance over the state-of-the-art (from $70.44\%$ to $75.14\%$ in absolute Dice scores). Thus, \ac{soars} provides a powerful and principled means to manage the highly complex segmentation space of \acp{oar}.
   
\end{abstract}

\acresetall
%%%%%%%%% BODY TEXT
\vspace{-1em}
\section{Introduction}
    \begin{figure}[ht]
    \centering
    \includegraphics[width=0.96\columnwidth]{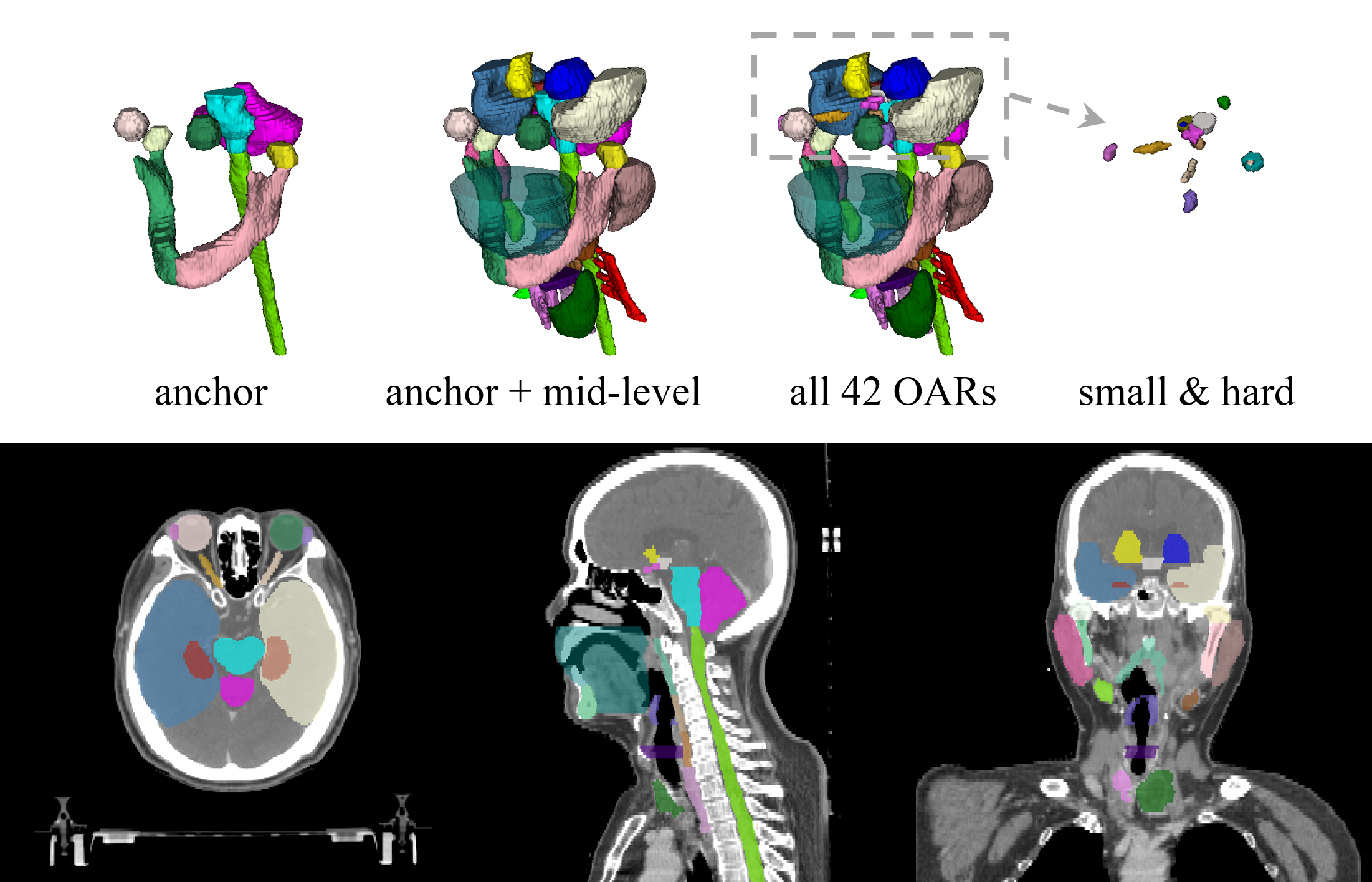}
    \caption{Illustration of 42 OARs in 3D demonstrating their various contrasts, sizes, and shapes in RTCT.}
    \label{fig:movtivation}
    \vspace{-1em}
    \end{figure}
    
    \Ac{hn} cancer is one of the most common cancers worldwide~\cite{jemal2011global}. High-precision radiation therapy, \eg  intensity-modulated radiotherapy, has been widely used for \ac{hn} cancer treatment because of its ability for highly conformal dose delivery. In this process, the radiation dose to normal anatomical structures, \ie \acp{oar}, should be controlled to minimize post-treatment complications~\cite{harari2010emphasizing}. This requires accurate delineation of tumors and \acp{oar} in \ac{RTCT} images~\cite{nikolov2018deep,cardenas2018auto,jin2019accurate,jin2019deep, lin2019deep, tang2019clinically}. Clinically, \ac{oar} segmentation is predominantly carried out manually by radiation oncologists. Manual delineation is not only time consuming, \eg $>$ 2 hrs for 9 \acp{oar}, but also suffers from large inter-practitioner variability~\cite{harari2010emphasizing}. Unsurprisingly, with more \acp{oar} included, time requirements increase significantly, limiting the number of patients who may receive timely radiotherapy~\cite{mikeljevic2004trends}. These issues have spurred efforts toward automatic \ac{oar} segmentation in \ac{hn} cancer~\cite{raudaschl2017evaluation}. Despite this progress, performance gaps remain, calling for approaches better tailored to this distinct and challenging problem. This is the goal of our work. 
    
    By their nature, \ac{hn} \acp{oar} are 1) complex in anatomical shapes, 2) dense in spatial distributions, 3) large in size variations, and 4) low in \ac{RTCT} image contrast.  Currently, deep \acp{cnn} are a dominant approach\cite{tong2018hierarchical, wang2017hierarchical,ibragimov2017segmentation, tong2018fully, gao2019focusnet, nikolov2018deep, tang2019clinically, zhu2019anatomynet, agn2019modality}. However, existing methods either perform whole volume segmentation~\cite{nikolov2018deep, zhu2019anatomynet} or segmentation-by-detection~\cite{gao2019focusnet, tang2019clinically}. Yet, model optimization becomes increasingly difficult as greater numbers of \acp{oar} need to be segmented. Leveraging insights from clinical practices can help ease the corresponding difficulties. 
       
    Within the clinic, radiation oncologists typically refer to easy \acp{oar} when delineating harder ones, \eg the eyes, brain stem, and mandible, to serve as anchors to segment hard \acp{oar}, such as different types of soft-tissue \ac{hn} glands~\cite{tamboli2011computed}.  \Fig~\ref{fig:movtivation} visually illustrates this stratification. As such, this process suggests that automated solutions could benefit from also stratifying \acp{oar}, both to create anchors and to create tailor-made analysis workflows for each stratification. Indeed, Gao~\etal~\cite{gao2019focusnet} showed that exploiting two branches for \ac{oar} segmentation boosts overall performance. However, large \acp{oar} did not serve as support to small \acp{oar} in that work. Moreover, the network architecture was manually crafted and fixed across \ac{oar} stratifications. Yet, given their highly distinct natures, different \acp{oar} likely require different network architectures for optimal performance. It is difficult to see how regular \acp{cnn} can meet these needs.
    
    Our work fills this gap by introducing \ac{soars}, a novel stratified learning framework to segment \acp{oar}. \ac{soars} divides \acp{oar} into three levels, \ie anchor, mid-level, and \ac{sh}. Emulating clinical practice, each is processed using tailored workflows. \textit{Anchor \acp{oar}} are high in intensity contrast and low in inter- and intra-reader variability. Thus these can be segmented first to provide informative location references to the harder categories. \textit{Mid-level \acp{oar}} are low in contrast, but not inordinately small. We provide anchor-level predictions as additional input for mid-level segmentation as guidance and reference-based grounding.  \textit{\Ac{sh} \acp{oar}} are very poor in contrast and very small. Similar to mid-level \acp{oar}, we use anchor \acp{oar} to guide \ac{sh} segmentation. However, we use a detection followed by segmentation strategy~\cite{gao2019focusnet}, to better manage the extremely unbalanced class distributions across the entire volume. While this workflow provides specialized frameworks for each \ac{oar} category, data processing could be even better tailored, as it is unlikely the same network architecture suits each stratification equally. Thus, we deploy an additional dimension of stratification, using \ac{nas} to automatically search the optimal architecture for each category. Concretely, we formulate the structure learning as a differentiable \ac{nas}~\cite{liu2018darts, liu2019auto, zhu2019v}, allowing for an automatic selection across 2D, 3D or \ac{p3d} convolutions with kernel sizes of 3 or 5 at each convolutional block. 
        
    Using four-fold cross-validation, we evaluate \ac{soars} on $142$ \ac{RTCT} images with $42$ annotated \acp{oar}, \textit{the most comprehensive \ac{hn} \ac{oar} dataset to date}. We demonstrate that both dimensions of our stratification, \ie category-specific processing and \ac{nas}, significantly impact performance. We achieve an average \ac{dsc} and \ac{hd} of $75.14\%$ and $6.98$mm, respectively, which corresponds to  improvements of $7.51\%$ and $2.41$mm, respectively over a non-stratified baseline. Compared to the state-of-the-art, a 3D Mask R-CNN based UaNet method~\cite{tang2019clinically}, we produce improvements of $4.70\%$ and $2.22$mm, in \ac{dsc} and \ac{hd}, respectively. Validation on a public dataset (the MICCAI 2015 \ac{oar} Challenge~\cite{raudaschl2017evaluation}), further confirms these compelling performance improvements. \guo{In summary, the contributions and novelty of this paper are three folds:
    \begin{itemize}
        \item Segmenting a comprehensive set of \acp{oar} is essential and critical for radiotherapy treatment planing in head and neck cancer. We work on the most clinically complete and desirable set of 42 \acp{oar} as compared to previous state-of-the-art work.
        \item Our main methodological contribution is the proposed whole framework on stratifying different organs into different categories of OARs which to be dealt respectively with tailored segmentors (achieved by \ac{nas}). Our method is a well-calibrated framework of integrating organ stratification, multi-stage segmentation and \ac{nas} in a synergy.
        \item Our idea of stratifying the 42 \acp{oar} into three levels comes from the combination of emulation of oncologists manual \ac{oar} contouring knowledge and the \ac{oar}'s size distributions.  To our best knowledge, this simple yet effective organ stratification scheme has not been studied for such a complex segmentation and parsing task like ours, by previous work.
    \end{itemize}
    }

%-------------------------------------------------------------------------
\section{Related Works}

\textbf{\acs{oar} Segmentation}
    There is a large body of work on  \acp{oar} segmentation. Atlas-based approaches~\cite{han2008atlas, isambert2008evaluation, saito2016joint, schreibmann2014multiatlas, sims2009pre, voet2011does} enjoy a prominent history~\cite{raudaschl2017evaluation,kosmin2019rapid}. Their main disadvantage is a reliance on accurate and efficient image registration~\cite{zhu2019anatomynet}, which is challenged by shape variations, normal tissue removal, abnormal tissue growth, and image acquisition differences~\cite{wu2019aar}. Registration often also take many minutes or even hours to complete. Another common approach is statistical shape or appearance models~\cite{cootes1995active, cootes2001active, rueckert2003automatic}. These have shown promise, but a prominent issue is that they can be limited to specific shapes described by the statistical model, which makes them less flexible when the number of \acp{oar} is large~\cite{fritscher2014automatic}. Of note is Tong \etal{}~\cite{tong2018hierarchical}, who applied intensity and texture-based fuzzy models using a hierarchical stratification.

    \begin{figure*}[ht]
    \centering
    \includegraphics[width=0.96\textwidth]{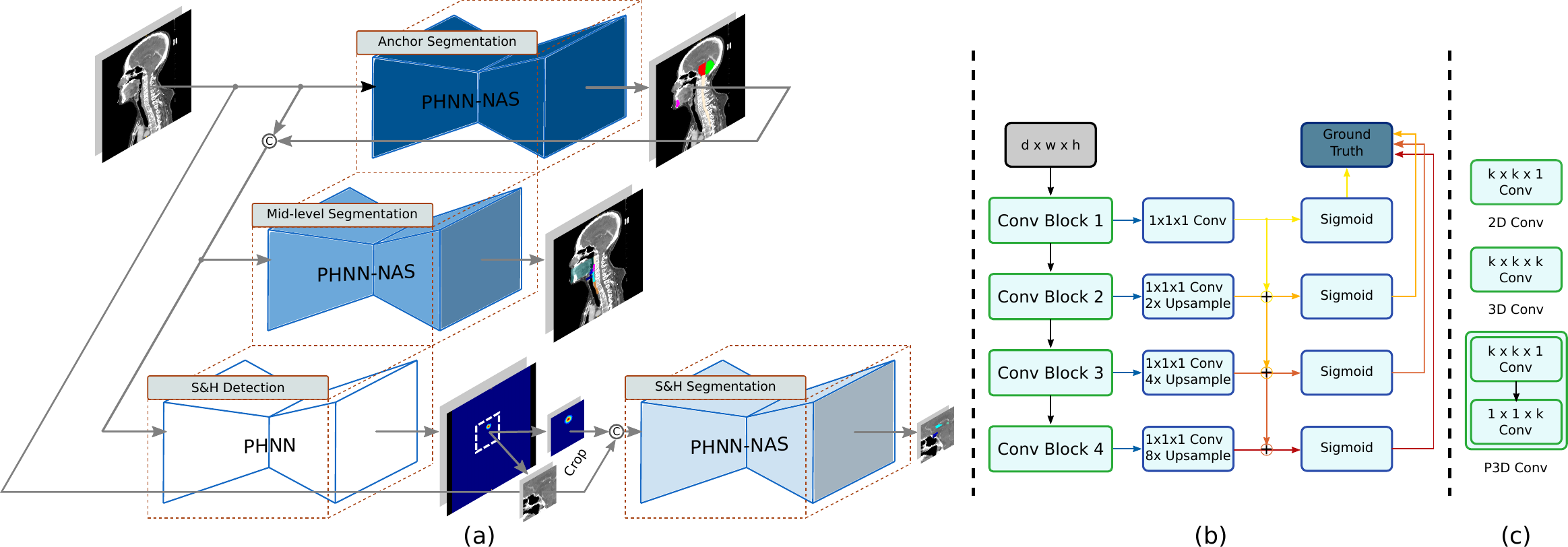}
    \caption{(a) \ac{soars} stratifies \ac{oar} segmentation across two dimensions: distinct processing frameworks and distinct architectures. We execute the latter using differentiable \ac{nas}. (b) depicts illustrates the backbone network (\ac{phnn}) with \ac{nas}, which allows for an automatic selection across 2D, 3D, \ac{p3d} convolutions. (c) demonstrates the \ac{nas} search space setting. }
    \label{fig: phnn-nas}
    \end{figure*}

Recently, deep \ac{cnn} based approaches have proven capable of delivering substantially better performance. Apart from early efforts~\cite{ibragimov2017segmentation}, \acp{fcn} have quickly becomes the mainstream method~\cite{zhu2019anatomynet,nikolov2018deep,tong2018fully,jin2019accurate}. To address data imbalance issues when faced with \ac{sh} \acp{oar}, FocusNet~\cite{gao2019focusnet} and UaNet~\cite{tang2019clinically} adopt a segmentation-by-detection strategy to achieve better segmentation accuracy. However, both approaches do not stratify \acp{oar}, and hence, cannot use easier \acp{oar} as support to more difficult ones.  Moreover, when the number of \acp{oar} is large, \eg $> 40$, optimization becomes more difficult. Finally, their network architecture remains manually fixed, which is less optimized for the distinct \acp{oar} categories.\\
\textbf{Stratified Learning}
    Stratification is an effective strategy to decouple a complicated task into easier sub-tasks.  Computer vision has a long history using this strategy. Several contextual learning models have been used to assist general object detection~\cite{heitz2008learning, rabinovich2007objects} within the conditional random field framework~\cite{lafferty2001conditional}. Instance learning, \ie instance localization, segmentation and categorization, often stratifies the problem into multiple sub-tasks~\cite{dai2016instance, he2017mask, chen2019hybrid}. Within medical imaging, stratified statistical learning has also been used to recognize whether a candidate nodule connects to any other major lung anatomies~\cite{wu2010stratified}. Yet, the use of stratified learning for semantic segmentation, particularly in the deep-learning era, is still relatively understudied in medical imaging. Within \ac{oar} segmentation, Tong \etal{}~\cite{tong2018hierarchical} have applied a hierarchical stratification, but this used a non-deep fuzzy-connectedness model. We are the first to execute stratified learning for deep \ac{oar} segmentation. \\
   \textbf{Neural Architecture Search}
    This is the process of automatically discovering better network architectures. Many \ac{nas} methods exploit reinforcement learning~\cite{zoph2018learning} or evolutionary algorithms~\cite{real2019regularized}.  However, both strategies are extremely computationally demanding. Differentiable \ac{nas}~\cite{liu2018darts, liu2019auto, zhu2019v} realize all candidate architectures simultaneously during optimization, limiting the allowable or feasible search spaces. Nonetheless, these approaches are highly practical means to tailor architectures. In this work, we follow the differentiable \ac{nas} formulation~\cite{liu2018darts, zhu2019v} to search the architectures for each of the three \ac{oar} stratifications. We explore the optimal kernel size and combination of 2D, 3D, and \ac{p3d} configurations. As such, we are the first to apply \ac{nas} to \ac{oar} segmentation.

\section{Methods}
    \Fig~\ref{fig: phnn-nas} depicts the \ac{soars} framework, which uses three processing branches to stratify anchor, mid-level and \ac{sh} \acp{oar} segmentation. A first stratification dimension is distinct processing frameworks.  \ac{soars} first segments the anchor \acp{oar}. Then, with the help of predicted anchors, mid-level and  \ac{sh} \acp{oar} are segmented. For the most difficult category of \ac{sh}, we first detect  center locations and then zoom-in to segment the small \acp{oar}. The deeply-supervised 3D \ac{phnn}~\cite{harrison2017progressive} is adopted as the backbone for all three branches, which uses deep supervision to progressively propagate lower-level features to higher-levels ones using a parameter-less pathway. We opt for this backbone due to its good reported performance in other \ac{RTCT} works~\cite{jin2019accurate,jin2019deep}. A second dimension of stratification uses differentiable \ac{nas} to search distinct \ac{phnn} convolutional blocks for each \ac{oar} category.

\subsection{Processing Stratification}
\label{sec:stratification}
    As mentioned, \ac{soars} segments \acp{oar} using three distinct frameworks, where \acp{oar} are divided according to clinician recommendations (the details for our $42$ \ac{oar} dataset is reported in \Sec~\ref{sec:exp_dataset}). We denote the training data of $N$ data instances as $\mathbb{S}=\left\{ X_n, Y_n^{\mathrm{A}}, Y_n^{\mathrm{M}}, Y_n^{\mathrm{S}} \right\} _{n=1}^{N}$, where $X_n$, $Y_n^{\mathrm{A}}$, $Y_n^{\mathrm{M}}$, and $Y_n^{\mathrm{S}}$ denote the input \ac{RTCT} and ground-truth masks for anchor, mid-level, and \ac{sh} \acp{oar}, respectively.  Here, we drop $n$, when appropriate, for clarity. Throughout, we will abuse matrix/vector notation, using boldface to denote vector-valued volumes and use vector concatenation as an operation across all voxel locations. 
        
    \noindent\textbf{Anchor branch:}
    Assuming we have $C$ classes, \ac{soars} first uses the anchor branch to generate  \ac{oar} prediction maps for every voxel location, $j$, and every output class, $c$:
    \begin{align}
        \hat{Y}^{\mathrm{A}}_c(j) &= p^{\mathrm{A}}\left( Y^{\mathrm{A}}(j) = c\, |\, X ; \mathbf{W}^{\mathrm{A}}\right) \label{eq:anchor} \mathrm{,} \\
        \hat{\mathbf{Y}}^{\mathrm{A}}&=\left[ \hat{Y}^{\mathrm{A}}_1\ldots\hat{Y}^{\mathrm{A}}_C \right] \mathrm{,} \label{eqn:anchor_vector}
    \end{align}
    where $p^{\mathrm{A}}(\cdot)$ and $\hat{\mathbf{Y}}^{\mathrm{A}}_c$ denote the \ac{cnn} functions and output segmentation maps, respectively. Here, predictions are vector valued 3D masks as they provide a pseudo-probability for every class. $\mathbf{W}^{(\cdot)}$ represents the corresponding \ac{cnn} parameters.
      
    Anchor \acp{oar} have high contrast compared to surrounding tissue or are in easy-to-locate regions; hence, it is relatively easy  to segment them directly and robustly based on pure appearance and context features. Consequently, they are ideal candidates to support the segmentation of other \acp{oar}.

    \noindent\textbf{Mid-level branch:}
    Most mid-level \acp{oar} are primarily soft tissue, which have low contrast and can be easily confused with other structures with similar intensities and shapes. Direct segmentation can lead to false-positives or over/under-segmentations. This can be addressed by using processing stratification to directly incorporate anchor predictions into mid-level learning, since the anchor predictions are robust and provide highly informative location and semantically-based cues. As demonstrated in \Fig~\ref{fig: phnn-nas}, we combine the anchor predictions with the \ac{RTCT} to create a multi-channel input: $\left[ X, \,\, \hat{\mathbf{Y}}^A\right]$:
    \begin{equation}
        \hat{Y}^{\mathrm{M}}_c(j) = p^{\mathrm{M}}\left( Y^{\mathrm{M}}(j) = c \, | \, X, \hat{\mathbf{Y}}^{\mathrm{A}}; \mathbf{W}^{\mathrm{M}}\right) \mathrm{.}
    \end{equation}
    In this way, the mid-level branch leverages both the \ac{CT} intensities as well as the anchor \ac{oar} guidance, which can be particularly helpful in managing regions with otherwise similar \ac{CT} appearance. Like \eqref{eqn:anchor_vector}, we can collect mid-level predictions into a vector-valued entity $\hat{\mathbf{Y}}^M$.
    
    \noindent\textbf{Small \& hard branch:} 
    In this branch, we further decouple segmentation into a detection followed by segmentation process. Directly segmenting the fine boundaries of \ac{sh} \acp{oar} from \ac{CT} is very challenging due to the poor contrast and the extremely imbalanced foreground and background distributions when considering the entire volume. In contrast, the detection of center regions of \ac{sh} \acp{oar} is a much easier problem, since the \ac{hn} region has relatively stable anatomical spatial distribution. This means that the \textit{rough} locations of \ac{sh} \acp{oar} can be inferred from the \ac{CT} context with confidence. Once the center location is detected, a localized region can be cropped out to focus on segmenting the fine boundaries in a zoom-in fashion. This has similarities to Gao \etal's~\cite{gao2019focusnet} approach to segment small organs. 
    
    For detecting \ac{sh} \ac{oar} centers, we adopt a simple yet effective heat map regression method~\cite{wei2016convolutional, xu2018less}, where the heat map labels are created at each organ center using a 3D Gaussian kernel. Similar to the mid-level branch, to increase detection robustness and accuracy we also combine the anchor branch predictions with the \ac{RTCT} as the detection input channels:

    \begin{align}
        \hat{\mathbf{H}} = f(X,\hat{\mathbf{Y}}^{A};\mathbf{W}^{D})
        \label{eq:small-det}
    \end{align}
    where $\hat{\mathbf{H}}$ denotes the predicted heat maps for every \ac{sh} \ac{oar}. Like the segmentation networks, we use the same \ac{phnn} backbone for $f(\cdot)$. Given the resulting regressed heat map, we choose the pixel location corresponding to the highest value, and crop a \ac{voi} using three-times the extent of the maximum size of the \ac{oar} of interest. With the \ac{voi} cropped, \ac{soars} can then segment fine boundaries of the \ac{sh} \acp{oar}. As illustrated in \Fig~\ref{fig: phnn-nas}, we concatenate the output from \Eq~(\ref{eq:small-det}) together with the cropped \ac{RTCT} image as the input to the \ac{sh} \ac{oar} segmentation network:
    \begin{align}
        \hat{Y}^{\mathrm{S}}_{c}(j) = p^{\mathrm{S}}\left( {Y}^{\mathrm{S}}(j) = c \, | \, X, \hat{\mathbf{H'}} ; \mathbf{W}^{\mathrm{S}}\right) \mathrm{,} \label{eqn:sh}
    \end{align}
    where here it's understood that \eqref{eqn:sh} is only operating on the cropped region.

\subsection{Architectural Stratification}
    While stratifying \acp{oar} into different processing frameworks with distinct inputs and  philosophies is key to pushing performance, more can be done. Namely, considering the significant variations in \ac{oar} appearance, shape, and size, it is likely that each \ac{oar} type would benefit from segmentation branch architectures tailored to their needs. To do this, \ac{soars} automatically searches network architectures for each branch, adding an additional dimension to the stratification. Throughout, we use  \ac{phnn}~\cite{harrison2017progressive} as the base backbone. The whole network structure is illustrated in \Fig~\ref{fig: phnn-nas}, in which the architecture is learned in a differentiable way~\cite{liu2018darts}. 
    
    Let $\phi(\, \cdot \, ; \boldsymbol \omega_{x\times y \times z})$ denote a composite function of the following consecutive operations: batch normalization, a rectified linear unit and a convolution with an $x\times y \times z$ dimension kernel. If one of the dimensions of the kernel is set to $1$, it reduces to a 2D kernel. As shown in \Eq~(\ref{eq:nas_kernels}), we search a set of possible architectures that include: 2D convolutions, 3D convolutions, or pseudo-3D convolution with either kernel sizes of 3 or 5:
    \begin{align}
    \begin{split}
        \phi_{\text{2D}_3} = \phi\left( \, \cdot \, ; \boldsymbol \omega_{3\times3\times1} \right) \mathrm{,}\\
        \phi_{\text{2D}_5} = \phi\left( \, \cdot \, ; \boldsymbol \omega_{5\times5\times1} \right) \mathrm{,}\\
        \phi_{\text{3D}_3} = \phi\left( \, \cdot \, ; \boldsymbol \omega_{3\times3\times3} \right) \mathrm{,}\\
        \phi_{\text{3D}_5} = \phi\left( \, \cdot \, ; \boldsymbol \omega_{5\times5\times5} \right) \mathrm{,}\\
        \phi_{\text{P3D}_3} = \phi\left( \phi\left( \, \cdot \, ; \boldsymbol \omega_{3\times3\times1} \right);\boldsymbol \omega_{1\times1\times3} \right) \mathrm{,}\\
        \phi_{\text{P3D}_5} = \phi\left( \phi\left( \, \cdot \, ; \boldsymbol \omega_{5\times5\times1} \right);\boldsymbol \omega_{1\times1\times5} \right) \mathrm{,}\\
        \boldsymbol \Phi = \left\{ \phi_{\text{2D}_3}, \phi_{\text{2D}_5}, \phi_{\text{3D}_3}, \phi_{\text{3D}_5}, \phi_{\text{P3D}_3}, \phi_{\text{P3D}_5} \right\} \mathrm{,}
    \end{split}
    \label{eq:nas_kernels}
    \end{align}
    where $\boldsymbol \Phi$ denotes the search space of possible architectures. For simplicity, instead of a layer-by-layer architecture search, we use only one type of convolutional kernel to build each \ac{phnn} convolutional block. 
    
    Similar to~\cite{liu2018darts, zhu2019v}, we make the search space continuous by relaxing the categorical choice of a particular operation to a softmax over all 6 possible operations. More formally, if we index each possibility in \eqref{eq:nas_kernels} by $k$, then we can define a set of 6 learnable logits for each, denoted $a_{k}$. A softmax can then be used to aggregate all possible architectures into one combined output, $\phi'$:
    \begin{align}
        \gamma_{k} &= \dfrac{\text{exp}\left( \alpha_{k} \right)}{\sum_{m}\text{exp}\left( \alpha_{m} \right)} \\
        \phi' &= \sum_{k} \gamma_{k}\phi_{k} \mathrm{,}
    \label{eq:nas_search_alpha}
    \end{align}
 where we have dropped dependence on the input images for convenience. As Zhu \etal  demonstrated~\cite{zhu2019v}, this type of \ac{nas} scheme can produce significant gains within medical image segmentation. This creates a sort of super network that comprises all possible manifestations of \eqref{eq:nas_kernels}. This super network can be optimized in the same manner as standard networks. At the end of the \ac{nas}, the chosen network architecture of each block, $\tilde{\phi}$, can be determined by selecting the $\phi$ corresponding to the largest $\alpha_{k}$ value. If the index to this maximum is denoted $\tilde{k}$, then $\tilde{\phi} = \phi_{\tilde{k}}$. If we have $b$ blocks, then based on \eqref{eq:nas_search_alpha}, the searched network can be represented as $\tilde{p} \left(\, \cdot \,; \tilde{\mathbf{W}} \right) = \tilde{\phi}^b \left( \tilde{\phi}^{b-1} \left(\, \cdots \, \tilde{\phi}^1 \left(\, \cdot \,; \boldsymbol{\tilde{\omega}}^1 \right); \boldsymbol{\tilde{\omega}}^{b-1} \right) ; \boldsymbol{\tilde{\omega}}^b\right)$, where $\tilde{(\cdot)}$ denotes the searched network architecture. For consistency, we use the same strategy to search the network architecture for each branch of \ac{soars}. 
        
    \begin{table*}[]
    \centering
    \scalebox{.85}{
    \begin{tabular}{lcccclcccclccc}
    \multicolumn{4}{c}{Anchor OARs}   &  & \multicolumn{4}{c}{Mid-level OARs}   &  & \multicolumn{4}{c}{S\&H OARs} \\  \\
    \multicolumn{1}{c}{} & DSC   & HD   & ASD  &  &      & DSC   & HD    & ASD  &  &         & DSC   & HD   & ASD  \\ \cline{1-4} \cline{6-9} \cline{11-14} 
    Baseline    & 84.02 & 5.98 & 0.82 &  & Baseline      & 63.68 & 12.97 & 3.48 &  & Baseline         & 60.97 & 4.86 & 0.98 \\ \cline{1-4} \cline{6-9} \cline{11-14} 
    CT Only     & 84.14 & 5.25 & 0.79 &  & CT Only       & 67.31 & 12.03 & 3.97 &  & CT Only & 62.09 & 4.19 & 1.06 \\
    CT+NAS      & 85.73 & 4.77 & 0.77 &  & CT+Anchor     & 70.73 & 10.34 & 1.67 &  & CT+Heat map      & 71.75 & 2.93 & 0.52 \\ \cline{1-4}
       &       &      &      &  & CT+Anchor+NAS & 72.55 & 9.05  & 1.31 &  & CT+Heat map +NAS & 72.57 & 2.94 & 0.49 \\ \cline{6-9} \cline{11-14} 
    \end{tabular}
    }
    \vspace{0.5em}
    \caption{Quantitative results of the ablation studies of the proposed method using 1 fold of the dataset. The baseline network is a 3D \ac{phnn}. For \ac{sh} \acp{oar}, all methods, except the baseline, segment on \textit{predicted} \acp{voi}. The performance is measured by \ac{dsc} (unit: \%), \ac{hd} (unit: mm), and \ac{asd} (unit: mm).}
    \label{tab:ablation_studies}
    \end{table*}
    
    \begin{table}[]
    \center
    \scalebox{0.85}{
    \begin{tabular}{lcc}
    
     &       Dist (mm) \\ \hline
    CT Only   &   3.25$\pm$2.34 \\
    CT+Anchor &   2.91$\pm$1.74 \\ \hline
    \end{tabular}
    }
    \vspace{0.5em}
    \caption{\Ac{sh} \ac{oar} detection results measuring the average distance between regressed and true center points. }
    \label{tab:sh_detection}
    \vspace{-1em}
    \end{table}
    
    \begin{figure*}[h]
    \centering
    \includegraphics[width=0.94\textwidth]{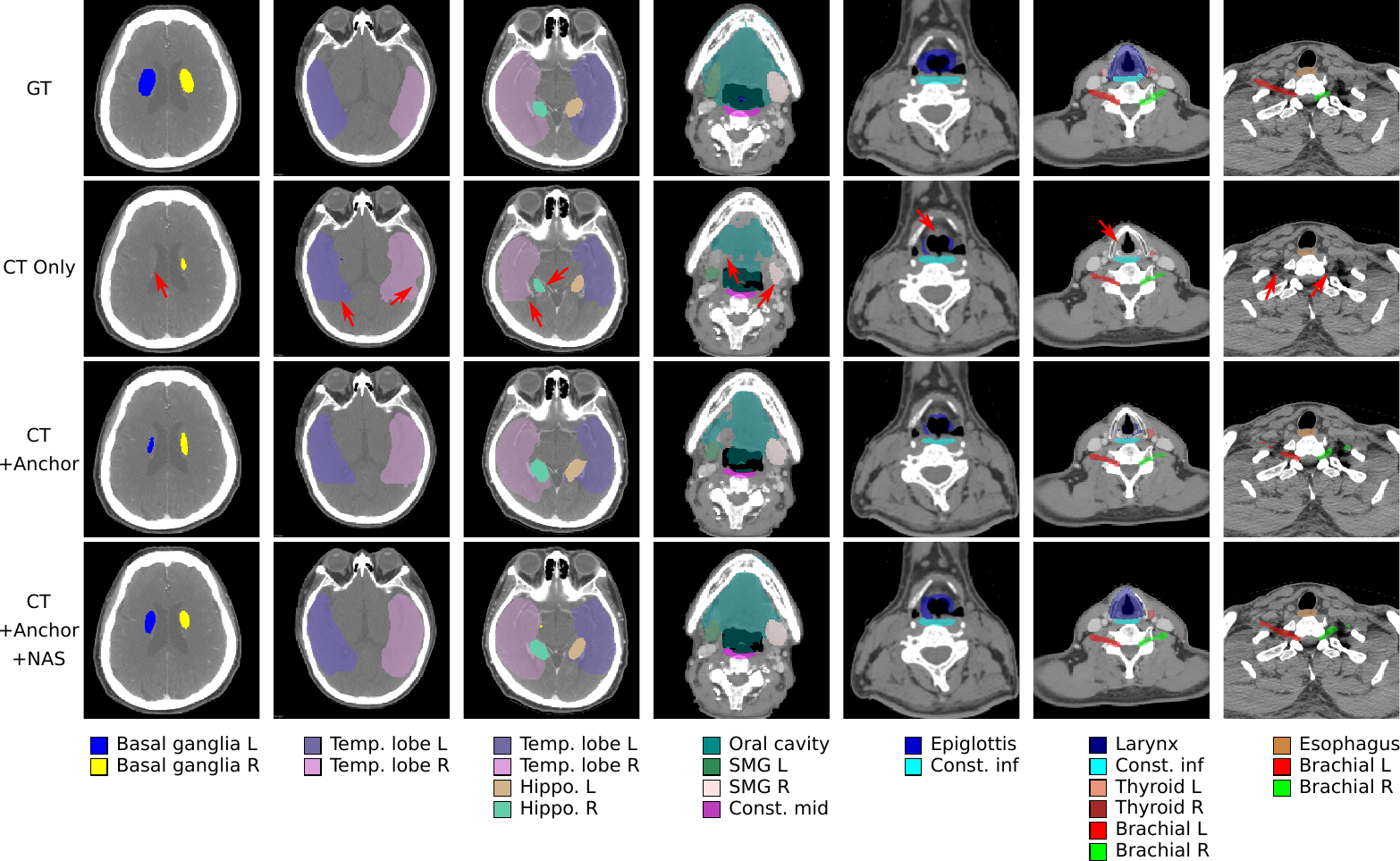}
    \caption{Qualitative mid-level \ac{oar} segmentation using different setups. The seven columns are seven representative axial slices in the \ac{RTCT} image.  For better comparison, we use red arrows to indicate the improvements. The $1^{st}$ row is the \ac{RTCT} image with \ac{oar} delineations of a radiation oncologist. The $3^{rd}$ row shows the impact of using anchor \acp{oar}, which can help the segmentation of soft-tissue mid-level \acp{oar}. The $4^{th}$ demonstrates the impact of \ac{nas}, indicating the necessity of adapting network architectures for different \acp{oar}.}
    \label{fig: result_quali}
    \vspace{-1em}
    \end{figure*}
    
\section{Experiments}
\subsection{Datasets and Preprocessing}
\label{sec:exp_dataset}
    To evaluate performance, we collected $142$ anonymized non-contrast \ac{RTCT} images in \ac{hn} cancer patients, where 42 \acp{oar} are delineated during the target contouring process for radiotherapy (hereafter denoted as \ac{hn} 42 dataset). Extensive 4-fold cross validation, split at the patient level, was conducted on the \ac{hn} 42 dataset to report results. We compare against other state-of-the-art methods including P-HNN~\cite{harrison2017progressive}, UNet~\cite{cciccek20163d}, and UaNet~\cite{tang2019clinically}.  To evaluate the effectiveness of \ac{soars}, we conducted two ablation studies using 1 fold of the dataset. Furthermore, we examined our performance using the public MICCAI 2015 head and neck auto-segmentation challenge data\footnote{\url{http://www.imagenglab.com/wiki/mediawiki/index.php?title=2015_MICCAI_Challenge}} (referred hereafter as MICCAI2015). This external testing set contains 9 \acp{oar} with 15 test cases. {\bf Evaluation metrics:} We report the segmentation performance using \ac{dsc} in percentage, \ac{hd} and \ac{asd} in mm. Note that we use \ac{hd} metric instead of \ac{hd}95 as reported in some previous works.
    
    {\bf \ac{hn} 42 \acp{oar} dataset:} Each CT scan is accompanied by 42 \ac{oar} 3D masks annotated by an experienced oncologist. The average CT size is $512\times512\times360$ voxels with an average resolution of $0.95\times0.95\times1.9$ mm. The specific \acp{oar} stratification is as follows. {\bf Anchor \acp{oar}}: brain stem, cerebellum, eye (left and right), mandible (left and right), spinal cord and temporomandibular joint (left and right). {\bf Mid-level \acp{oar}}: brachial plexus (left and right), basal ganglia (left and right), constrictor muscle (inferior, middle and superior), epiglottis, esophagus, hippocampus (left and right), larynx core, oralcavity, parotid (left and right), submandibular gland (left and right), temporal lobe (left and right), thyroid (left and right). {\bf \ac{sh} \acp{oar}}: cochlea (left and right), hypothalamus, inner ear (left and right), lacrimal gland (left and right), optic nerve (left and right), optic chiasm, pineal gland, and pituitary. 

    {\bf MICCAI2015 dataset:} This dataset has been extensively used by researchers to evaluate atlas and deep learning based \ac{hn} \ac{oar} segmentation.  It contains 33 training cases and 15 test cases with 9 \acp{oar} annotated. The 9 \acp{oar} include brain stem, mandible, optic chiasm, optic nerve (left and right), parotid (left and right) and submandibular gland (left and right).

    {\bf Image preprocessing:} We apply a windowing of [-500, 1000] HU to every CT scan covering the intensity range of our target \acp{oar}, from which we extract $128\times128\times64$ \acp{voi} as training samples for the anchor and mid-level branches as well as the detection module in the \ac{sh} branch. The heat map labels in the detection module is a 3D Gaussian distribution with a standard deviation of 8mm. The training \acp{voi} are sampled in two manners: (1) we randomly extract \acp{voi} centered within each of the \acp{oar} to ensure sufficient positive samples. (2) we randomly sample additional 15 \acp{voi} from the whole volume to obtain sufficient negative examples. This results in on average 70 \acp{voi} per \ac{CT} scan. We further augment the training data by applying random scaling between $0.8-1.2$. In testing, 3D sliding windows with sub-volumes of $128\times128\times64$ and strides of $96\times96\times32$ voxels are used. The probability maps of sub-volumes are aggregated to obtain the whole volume prediction, taking on average $20$s to process one input volume using a single GPU. 

\subsection{Implementation Details}
\label{sec:exp_implement}
    We implemented \ac{soars} in PyTorch\footnote{\url{https://pytorch.org/}}, and trained it on an NVIDIA Quadro RTX 8000.  The RAdam solver~\cite{liu2019variance} is used to optimize all models with a momentum of $0.9$ and a weight decay of $0.005$. The \ac{dsc} loss is used for the segmentation task training. The \ac{sh} detection branch is trained using L2 loss with a $0.01$ learning rate.  

    We exploit \ac{nas} to search the optimal network architecture for each branch. For the \ac{nas} parameter $\alpha_{k}$, we first fix $\alpha_{k}$ for $20$ epochs. Then we update $\alpha_{k}$ and the network weights for an additional $30$ epochs. The batch size for \ac{nas} training is set to $2$. Note that we use only the validation set for $\boldsymbol{\alpha}$ updating. The ratio between the training set and the validation set is 2:1. The initial learning rate is set to $0.005$ for the anchor and mid-level branches, and $0.001$ for the \ac{sh} branch.  
    
    After \ac{nas} is completed, we retrain the searched network from scratch with a batch size of $12$. The batch size is set to be $12$. The initial learning rate is set to $0.01$ for the anchor and mid-level branches, and $0.005$ for the \ac{sh} branch. The detailed training strategy is described as follows: 1) We train the anchor branch for $50$ epochs; 2) We fix the parameters of the anchor branch and concatenate its output to the original RTCT, followed by further training the mid-level and S\&H branches for $50$ epochs; 3) Finally we fine-tune the whole framework in an end-to-end manner for $10$ epochs.

\subsection{Processing Stratification}
\label{sec:exp_oar}
    We first evaluate the effectiveness of the processing stratification of \ac{soars}. The ablation results for segmenting the anchor, mid-level and \ac{sh} \acp{oar} are shown in Table~\ref{tab:ablation_studies}. The baseline comparison is the 3D \ac{phnn} model trained on all 42 \acp{oar} together. When anchor \acp{oar} are stratified to train only on themselves, there is a slight improvement as compared to the baseline model, consistent with the observation that anchor \acp{oar} generally have good contrast and are easy to optimize.  However, when focusing on mid-level \acp{oar}, there is a marked \ac{dsc} score improvement ($3.63\%$) when only training on mid-level \acp{oar} instead of training on all. This demonstrates the difficulty in segmenting a large number of organs together without considering their differences. When further adding anchor \ac{oar} predictions as support, both \ac{dsc} scores and the ASD experience large improvements, \ie from $67.31\%$ to $70.73\%$ in \ac{dsc} and $3.97$ to $1.67$mm in ASD. These significant error reductions indicate that anchor \acp{oar} serve as effective references to better delineate the hard-to-discern boundaries of mid-level organs (most are soft-tissue). \Fig~\ref{fig: result_quali} depicts qualitative examples of segmenting mid-level \acp{oar}. As can be seen, our method achieves much better visual results.
    
\begin{figure}
    \centering
    \includegraphics[width=0.9\columnwidth]{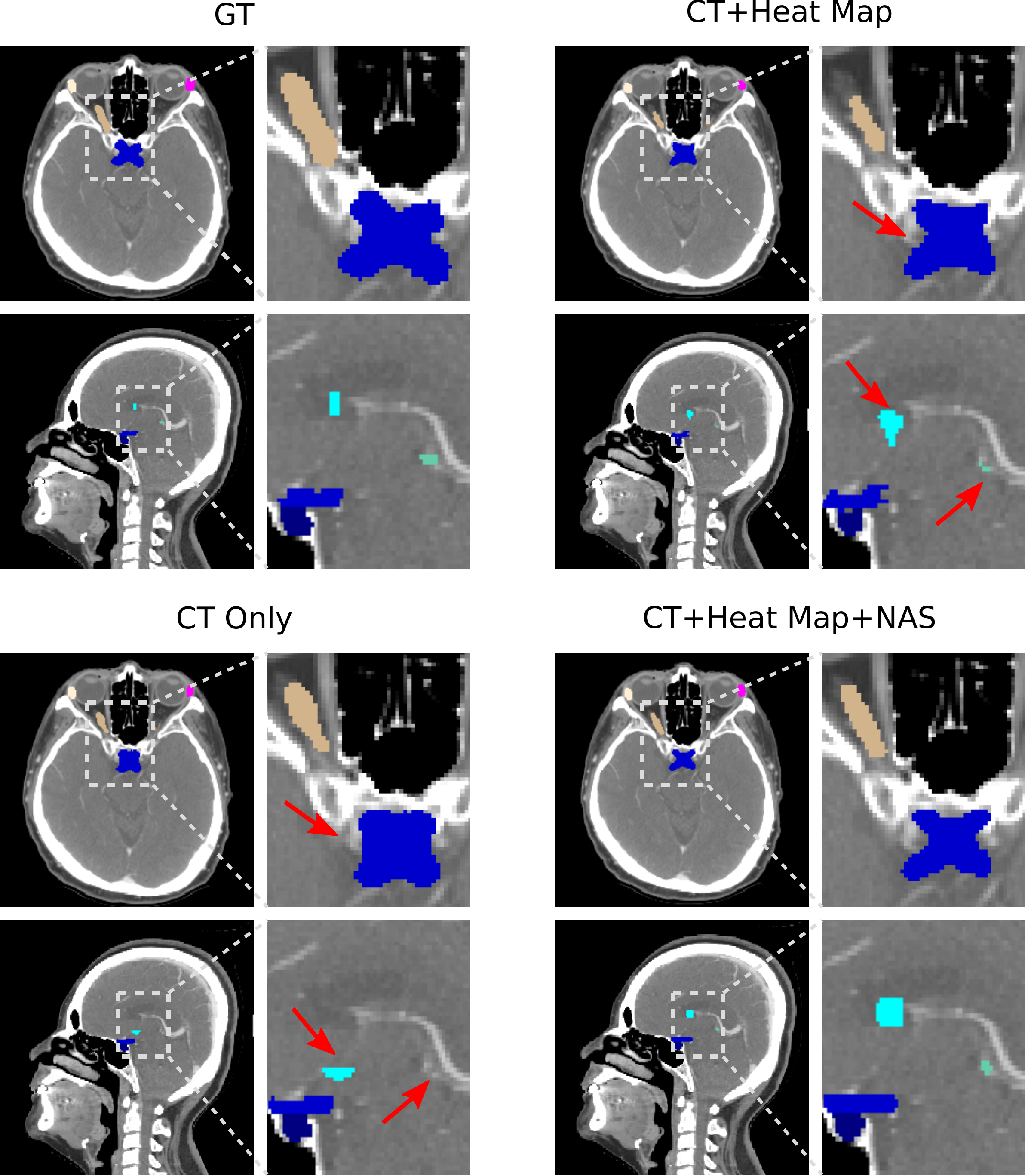}
    \caption{Examples of \ac{sh} \ac{oar} segmentation using different setups. For visualiztion purpose, the dashed rectangles are enlarged for highlighting improvements. As indicated using the red arrows, the proposed method achieves visually better optic chiasm, hypothalamus, and pineal gland segmentation.}
    \label{fig:result_quali_sh}
    \vspace{-1em}
    \end{figure}

    For the \ac{sh} branch, we first report the accuracy of the regressed center-point using the detection-by-segmentation network. As \Tab~\ref{tab:sh_detection} demonstrates, the center points of \ac{sh} \acp{oar} can be detected with high robustness. Moreover, when using the anchor \acp{oar} as support, the distance errors between regressed and true center points are further reduced. In our experiments, no \ac{sh} \ac{oar} was missed by our detection-by-segmentation strategy, demonstrating the robustness of our approach. Now focusing on the segmentation results of \Tab~\ref{tab:ablation_studies}, by cropping the \ac{voi} using the detection module, there is remarkable improvement in segmenting the \ac{sh} \acp{oar}, moving \ac{dsc} from $62.09\%$ to $71.75\%$, as compared against  directly segmenting from the CT. This further demonstrates the value of our processing-based stratification method, which provides for optimal treatment of \ac{oar} categories with different characteristics. As the examples of \Fig~\ref{fig:result_quali_sh} demonstrate, the benefits of processing stratification for \ac{sh} \acp{oar} is clearly shown in the optic chiasm, hypothalamus, and pineal gland, which are insufficiently segmented/missed when using only \ac{RTCT} for prediction. 
    
    \begin{table*}[]
    \centering
    \scalebox{0.85}{
    \begin{tabular}{lccclccclccclccc}
        & \multicolumn{3}{c}{Anchor OARs} &  & \multicolumn{3}{c}{Mid-level OARs} &  & \multicolumn{3}{c}{S \& H OARs} &  & \multicolumn{3}{c}{All OARs} \\ \cline{2-4} \cline{6-8} \cline{10-12} \cline{14-16} 
        & DSC       & HD       & ASD      &  & DSC        & HD         & ASD      &  & DSC       & HD       & ASD      &  & DSC      & HD       & ASD    \\ \hline
    UNet~\cite{cciccek20163d}    & 82.97     & 8.90     & 1.06     &  & 63.61      & 11.06      & 1.92     &  & 59.64     & 6.38     & 1.31     &  & 66.62    & 9.26     & 1.86   \\
    \ac{phnn}~\cite{harrison2017progressive} & 84.26     & 6.12     & 1.18     &  & 65.19      & 13.15      & 2.97     &  & 59.42     & 5.23     & 0.82     &  & 67.62    & 9.39    & 2.23   \\
    UaNet~\cite{tang2019clinically}         & 84.30     & 8.89     & 1.72     &  & 69.40      & 11.57      & 2.06     &  & 61.85     & 5.28     & 1.53     &  & 70.44    & 9.20     & 1.83   \\ \hline
    SOARS & \textbf{85.04}     & \textbf{5.08}     & \textbf{0.98}     &  & \textbf{72.75}      & \textbf{10.10}      & \textbf{1.66}     &  & \textbf{71.90}     & \textbf{2.93}     & \textbf{0.53}     &  & \textbf{75.14}    & \textbf{6.98}     & \textbf{1.12}   \\ \hline
    \end{tabular}
    }
    \vspace{0.5em}
    \caption{Quantitative results of different approaches on segmenting the 42 \ac{hn} \acp{oar} using the 4-fold cross validation. Our proposed SOARS achieves the best performance in all metrics (indicated in bold). }
    \label{tab:sota_comparison}
    \end{table*}
    
    \begin{table*}[]
    \centering
    \scalebox{0.81}{
    \begin{tabular}{lcccccccccc}
      & & & & \multicolumn{2}{c}{Optic Nerve}        & \multicolumn{2}{c}{Parotid}            & \multicolumn{2}{c}{SMG}     &     \\ \cline{5-10}
      & \multirow{-2}{*}{Brain Stem}         & \multirow{-2}{*}{Mandible}           & \multirow{-2}{*}{Optic Chiasm}       & Lt           & Rt           & Lt           & Rt           & Lt           & Rt           & \multirow{-2}{*}{All OARs} \\ \hline
    Ren~\etal~\cite{ren2018interleaved}    & -            & -            & 58.0$\pm$17.0           & 72.0$\pm$8.0 & 70.0$\pm$9.0 & -            & -            & -            & -             & -  \\
    Wang~\etal~\cite{wang2017hierarchical} & \textbf{90.0$\pm$4.0}   & 94.0$\pm$1.0 & -            & -            & -            & 83.0$\pm$6.0 & 83.0$\pm$6.0 & -            & -            & -  \\
    AnatomyNet~\cite{zhu2019anatomynet}     & 86.7$\pm$2.0 & 92.5$\pm$2.0 & 53.2$\pm$15.0           & 72.1$\pm$6.0 & 70.6$\pm$10.0           & 88.1$\pm$2.0 & 87.3$\pm$4.0 & 81.4$\pm$4.0 & 81.3$\pm$4.0 & 79.2          \\
    FocusNet~\cite{gao2019focusnet}        & 87.5$\pm$2.6 & 93.5$\pm$1.9 & 59.6$\pm$18.1           & 73.5$\pm$9.6 & \textit{74.4$\pm$7.}2 & 86.3$\pm$3.6 & \textit{87.9$\pm$3.1} & 79.8$\pm$8.1 & 80.1$\pm$6.1 & 80.3          \\
    UaNet~\cite{tang2019clinically}          & 87.5$\pm$2.5 & \textit{95.0$\pm$0.8} & \textit{61.5$\pm$10.2          } & \textit{74.8$\pm$7.1} & 72.3$\pm$5.9 & \textbf{88.7$\pm$1.9}   & 87.5$\pm$5.0 & \textit{82.3$\pm$5.2} & \textit{81.5$\pm$4.5}  & \textit{81.2}          \\ \hline
    SOARS          & \textit{87.6$\pm$2.8} & \textbf{95.1$\pm$1.1}   & \textbf{64.9$\pm$8.8}   & \textbf{75.3$\pm$7.1}   & \textbf{74.6$\pm$5.2}   & \textit{88.2$\pm$3.2} & \textbf{88.2$\pm$5.2}   & \textbf{84.2$\pm$7.3}   & \textbf{83.8$\pm$6.9}    & \textbf{82.4}          \\ \hline
    \end{tabular}
    }
    \vspace{0.5em}
    \caption{For MICCAI 2015 9 \acp{oar} segmentation challenge, the proposed method achieves 7 (in bold) best performance and 2 (in italic font) second best performance. }
    \label{tab:miccai_results}
    \vspace{-1em}
    \end{table*}

\subsection{Architectural Stratification}
\label{sec:exp_nas}
    \Tab~\ref{tab:ablation_studies} also outlines the performance improvements provided by \ac{nas}. As can be seen, all three branches trained with \ac{nas} consistently produce more accurate segmentation results than those trained with the baseline 3D \ac{phnn} network. This validates the effectiveness of \ac{nas} on complicated segmentation tasks. For the three branches, the anchor and mid-level branches have considerable performance improvement, from $84.14\%$ to $85.73\%$ and $70.73\%$ to $72.55\%$ in \ac{dsc} scores respectively, while the \ac{sh} branch provides a marginal improvement ($0.82\%$ in \ac{dsc} score). For segmenting the \ac{sh} \acp{oar}, the strong priors of detected heat maps may have already made the segmentation task much easier. Nonetheless, considering the dramatic improvements already provided by the stratified approach in \Sec~\ref{sec:exp_oar}, the fact that \ac{nas} is able to boost performance even further attests to its benefits. Some qualitative examples demonstrating the effectiveness of \ac{nas} are shown in \Fig~\ref{fig: result_quali} and \Fig~\ref{fig:result_quali_sh}. 
    
    The searched network architectures for the anchor branch are 2D-kernel3, 2D-kernel5, 2D-kernel3 and 3D-kernel5 for the four convolution blocks, while for the mid-level branch they are 2D-kernel3, 2.5D-kernel5, 2D-kernel3 and 2.5D-kernel5. This is an interesting result, as it indicates that 3D kernels may not always be the best choice for segmenting objects with reasonable size, as mixed 2D or \ac{p3d} kernels dominate both branches. Consequently, it is possible that much computation and memory used for 3D networks could be avoided by using an appropriately designed 2D or \ac{p3d} architecture. For the \ac{sh} branch, the search architecture is 2D-kernel3, 3D-kernel5, 2D-kernel3 and 3D-kernel5 for the four convolution blocks. As can be seen, more 3D kernels are used, consistent with the intuition that small objects with low contrast rely more on the 3D spatial information for better segmentation.
    
    Intuitively, it would be interesting to let the network search the \ac{oar} levels. However, \ac{nas} becomes computationally unaffordable since automatically stratifying anchor OARs alone is at the complexity of $\mathcal{C}_{42}^{9}\times$ more expensive.

\subsection{Comparison to State-of-the-art}
\label{sec:exp_base}
    \Tab~\ref{tab:sota_comparison} compares \ac{soars} against 3 \ac{sota} \ac{oar} segmentation methods, \ie UNet~\cite{cciccek20163d}, \ac{phnn}~\cite{harrison2017progressive}, and UaNet~\cite{tang2019clinically}, using the 4-fold cross-validation on the \ac{hn} 42 OARs dataset. We also tested  anatomyNet~\cite{zhu2019anatomynet}, but it consistently missed very small organs, so we do not report its results. Although \ac{phnn}~\cite{harrison2017progressive} achieves comparable performance on the anchor and \ac{sh} \acp{oar} segmentation with UaNet~\cite{tang2019clinically}, it has decreased performance for mid-level \acp{oar}. UaNet is a modified version of 3D Mask R-CNN~\cite{he2017mask}, which conducts object segmentation within the detected boxes. Hence, it decouples the whole complicated task into detection followed by segmentation, possibly accounting for the better segmentation accuracy for the mid-level \acp{oar} as compared to \ac{phnn}~\cite{harrison2017progressive}. Nonetheless, despite being much simpler, \ac{phnn} is still able to match or beat UaNet on the \ac{sh} \acp{oar}, demonstrating its effectiveness as a baseline and backbone method for \ac{soars}. When considering \ac{soars}, consistent improvements can be observed in all metrics as compared to all competitors, with $4.70\%$ absolute \ac{dsc} increases and $2.22mm$ \ac{hd} error reduction as compared to UaNet~\cite{tang2019clinically}.

\subsection{MICCAI2015 Challenge}
\label{sec:exp_miccai2015}
% substantial improvements on small \& hard \ac{oar} segmentation. 
    We use the MICCAI2015 dataset as an external dataset to further demonstrate the generalizability of \ac{soars}. Similar to other comparison methods, we trained our framework from scratch using the MICCAI2015 training set. We get an average \ac{dsc} of $82.4\%$, which has $1.2\%$ improvement as compared to~\cite{tang2019clinically}, or $2.1\%$ over~\cite{gao2019focusnet}. Compared to competitor methods, we achieve 7 best performance and 2 second best performance on all 9 \acp{oar}, especially the most difficult optic chiasm, where we have a $3.4\%$ improvement on \ac{dsc} as compared to the best previous result achieved by UaNet~\cite{tang2019clinically}. These results on the MICCAI2015 dataset further validate the effectiveness and consistency of our method, reinforcing its value.
    
\section{Conclusion}

    This work presented \ac{soars}, a novel framework that stratifies \ac{hn} \acp{oar} segmentation into two dimensions. Inspired by clinical practices, we stratify \acp{oar} into three categories of anchor, mid-level and \ac{sh}, providing customized processing frameworks for each. Importantly, the mid-level and \ac{sh} branches build off of the anchor branch's more reliable predictions. Additionally, we stratify network architectures, executing an effective \ac{nas} for each. We test on the most comprehensive \ac{hn} dataset to date that comprises 42 different \acp{oar}. Comparing to \ac{sota} methods, the improvements are most significant for the mid-level and \ac{sh} \acp{oar}.  With this, we demonstrate that our proposed \ac{soars} can outperform all state-of-the-art baseline networks, including the most recent representative work UaNet~\cite{tang2019clinically}, by margins as high as $4.70\%$ in \ac{dsc}. Thus, our work represents an important step forward toward reliable and automated \ac{hn} \ac{oar} segmentation.

{\small
\bibliographystyle{ieee_fullname}
\bibliography{egbib}
}

\clearpage
\section*{Supplementary Material}
    \subsection*{Performance of OAR segmentation}
    
\begin{figure*}[]
    \includegraphics[width=0.96\textwidth]{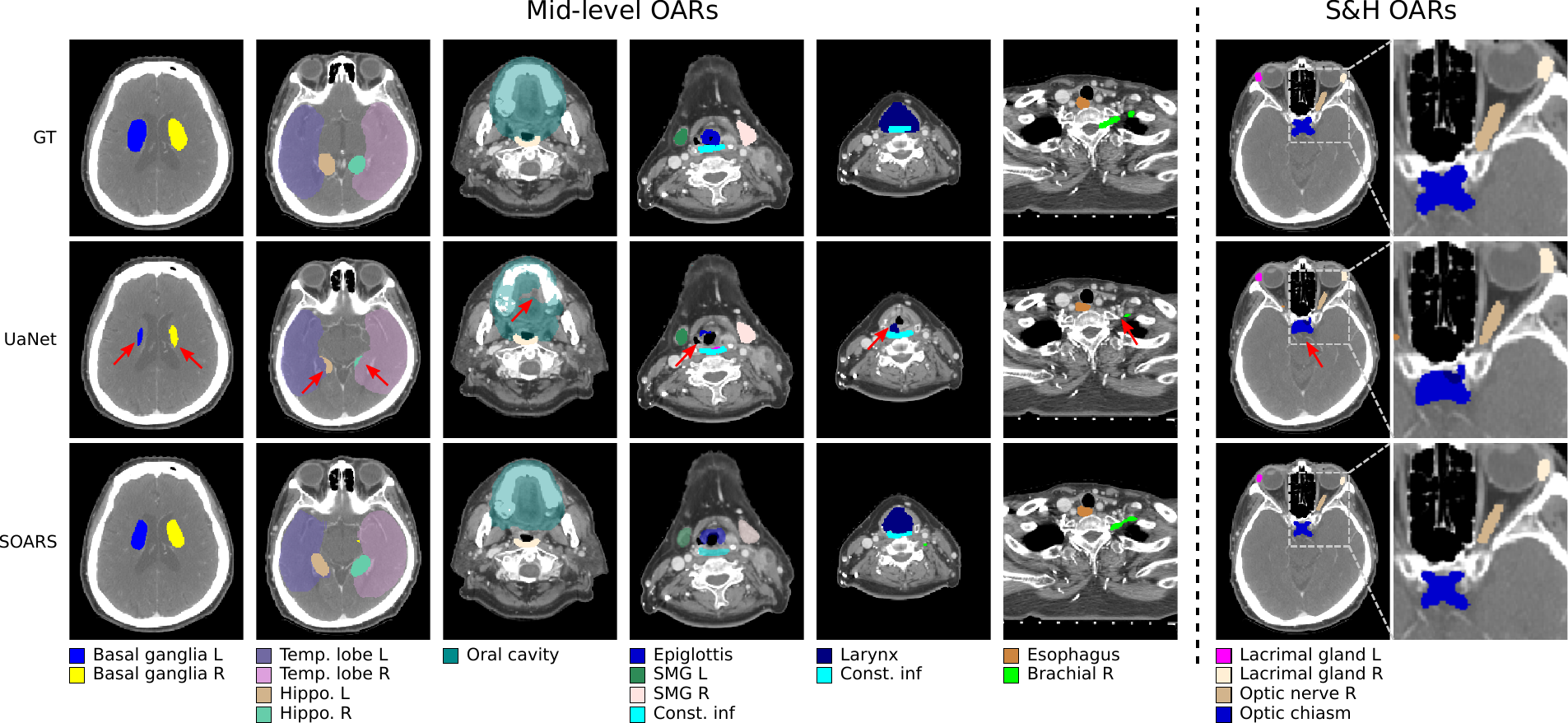}
    \caption{Qualitative illustration of the mid-level (left-hand side) and \ac{sh} (right-hand side) \ac{oar} segmentation using UaNet and the proposed \ac{soars}. The seven columns are seven representative axial slices in the \ac{RTCT} image. The $1^{st}$ column shows the \ac{oar} labels from a radiation oncologist, while the $2^{nd}$ and $3^{rd}$ columns are the predicted segmentation results by the UaNet and the proposed \ac{soars}, respectively.  For better comparison, we use red arrows to indicate the improvements. For visualization purpose, the dashed rectangles are enlarged for highlighting improvements on \ac{sh} \ac{oar} segmentation.}
    \label{fig: result_sup_quali}
    \end{figure*}

    In \Tab~\ref{tab:dice}, we report the category-by-category \ac{dsc} of the proposed SOARS against UNet~\cite{ronneberger2015u}, P-HNN~\cite{harrison2017progressive}, and UaNet~\cite{tang2019clinically}.  In \Tab~\ref{tab:hd}, we report the category-by-category \ac{hd} of the proposed SOARS against UNet, P-HNN, and UaNet. For both metrics, \ac{soars} achieved 30 out of 42 \acp{oar} best performance.  \ac{soars} performed slightly worse than UaNet on temporal lobe and temporomandibular joint segmentations in terms of \ac{dsc}. Yet, the \ac{dsc} differences are relatively small. We demonstrate some qualitative comparison results against UaNet in \Fig~\ref{fig: result_sup_quali}, where the improvements are indicated using red arrows. 
    
    \begin{table}[ht!]
    \centering
    \scalebox{0.7}{
    \begin{tabular}{lllll}
    \multicolumn{1}{l}{Organ} & \multicolumn{1}{c}{UNet} & \multicolumn{1}{c}{P-HNN} & \multicolumn{1}{c}{UaNet} & \multicolumn{1}{c}{SOARS} \\ \hline
    Basal Ganglia Lt          & \textbf{64.0$\pm$12.4}            & 63.5$\pm$16.6             & 63.6$\pm$13.7             & 63.8$\pm$13.7    \\
    Basal Ganglia Rt          & 64.7$\pm$13.9            & 63.5$\pm$14.2             & \textbf{67.4$\pm$15.0}    & 63.6$\pm$11.6             \\
    Brachial Lt               & 59.8$\pm$13.7            & 48.8$\pm$11.8             & 49.9$\pm$10.3             & \textbf{66.8$\pm$17.1}    \\
    Brachial Rt               & 58.8$\pm$13.7            & 49.4$\pm$7.0              & 53.5$\pm$8.0              & \textbf{65.5$\pm$14.2}    \\
    Brainstem                 & \textbf{81.7$\pm$5.4}    & 80.1$\pm$6.8              & 80.6$\pm$6.3              & 81.0$\pm$5.7              \\
    Cerebellum                & 83.2$\pm$2.7             & 88.8$\pm$2.8              & 90.1$\pm$2.8              & \textbf{90.2$\pm$2.3}     \\
    Cochlea Lt                & 64.0$\pm$17.6            & 67.2$\pm$10.4             & 66.5$\pm$12.6             & \textbf{72.3$\pm$12.2}    \\
    Cochlea Rt                & 64.2$\pm$10.0            & 67.2$\pm$10.4             & 68.2$\pm$12.6             & \textbf{69.5$\pm$12.4}    \\
    Const. inf                & 63.4$\pm$17.1            & 61.8$\pm$14.9             & \textbf{73.6$\pm$10.6}    & 65.0$\pm$18.3             \\
    Const. mid                & 64.9$\pm$15.4            & 63.1$\pm$14.5             & 66.1$\pm$11.3             & \textbf{66.9$\pm$15.1}    \\
    Const. sup                & 64.0$\pm$10.2            & 64.1$\pm$10.0             & 62.3$\pm$11.3             & \textbf{67.4$\pm$9.2}     \\
    Epiglottis                & 65.5$\pm$8.6             & 65.5$\pm$11.0             & 65.4$\pm$13.1             & \textbf{67.3$\pm$8.2}     \\
    Esophagus                 & 66.3$\pm$23.2            & 61.6$\pm$12.0             & \textbf{69.1$\pm$12.9}    & 67.0$\pm$14.0             \\
    Eye Lt                    & 83.4$\pm$7.4             & \textbf{86.4$\pm$3.4}              & 85.7$\pm$7.4              & \textbf{86.4$\pm$3.3}     \\
    Eye Rt                    & 82.7$\pm$6.3             & 85.9$\pm$3.3              & \textbf{86.7$\pm$4.3}     & 86.6$\pm$4.0              \\
    Hippocampus Lt            & 62.4$\pm$12.5            & 46.2$\pm$17.3             & 50.0$\pm$17.3             & \textbf{67.4$\pm$16.0}    \\
    Hippocampus Rt            & 62.2$\pm$14.3            & 45.2$\pm$12.1             & 52.2$\pm$17.6             & \textbf{67.9$\pm$18.9}    \\
    Hypothalamus              & 63.6$\pm$17.3            & 39.2$\pm$16.8             & 28.7$\pm$22.9             & \textbf{72.6$\pm$17.1}    \\
    Innerear Lt               & 62.4$\pm$12.1            & 58.4$\pm$10.6             & 68.8$\pm$10.9             & \textbf{78.8$\pm$8.1}     \\
    Innerear Rt               & 63.2$\pm$16.8            & 60.1$\pm$10.3             & 73.0$\pm$12.2             & \textbf{76.9$\pm$9.1}     \\
    Lacrimalgland Lt          & 59.2$\pm$10.5            & 54.7$\pm$11.5             & 64.1$\pm$16.0             & \textbf{70.7$\pm$8.0}     \\
    Lacrimalgland Rt          & 58.7$\pm$10.5            & 54.7$\pm$11.5             & 52.1$\pm$14.3             & \textbf{70.6$\pm$11.0}    \\
    Larynx core                    & 57.9$\pm$17.1            & 53.9$\pm$17.1             & 56.9$\pm$20.1             & \textbf{69.7$\pm$20.8}    \\
    Mandible Lt               & 87.4$\pm$2.9             & 90.2$\pm$2.0              & 88.2$\pm$12.1             & \textbf{91.7$\pm$1.8}     \\
    Mandible Rt               & 89.1$\pm$2.3             & 90.8$\pm$1.8              & 88.0$\pm$6.0              & \textbf{91.1$\pm$2.5}     \\
    Optic Chiasm              & 49.9$\pm$15.4            & 50.9$\pm$13.6             & 60.4$\pm$22.1             & \textbf{72.9$\pm$9.2}     \\
    Optic Nerve Lt            & 61.7$\pm$11.1            & 67.6$\pm$11.0             & 69.9$\pm$9.3              & \textbf{74.3$\pm$7.8}     \\
    Optic Nerve Rt            & 62.0$\pm$12.2            & 67.6$\pm$10.2             & 69.9$\pm$11.0             & \textbf{72.3$\pm$8.7}     \\
    Oralcavity                & 64.0$\pm$5.1             & 76.3$\pm$5.1              & 77.8$\pm$10.2             & \textbf{82.6$\pm$5.3}     \\
    Parotid Lt                & 64.7$\pm$5.8             & 78.2$\pm$5.1              & 82.8$\pm$6.2              & \textbf{84.5$\pm$4.2}     \\
    Parotid Rt                & 64.7$\pm$6.1             & 78.8$\pm$6.5              & 82.3$\pm$6.6              & \textbf{84.1$\pm$5.0}     \\
    Pineal Gland              & 46.4$\pm$29.3            & 60.2$\pm$16.5             & 63.6$\pm$26.4             & \textbf{70.4$\pm$14.7}    \\
    Pituitary                 & 60.4$\pm$11.0            & \textbf{65.2$\pm$11.0}    & 57.0$\pm$14.8             & 61.5$\pm$18.4             \\
    Spinalcord                & 83.5$\pm$6.2             & 83.7$\pm$3.6              & 82.7$\pm$7.4              & \textbf{84.6$\pm$2.4}     \\
    SMG Lt                    & 64.2$\pm$16.8            & 71.3$\pm$8.8              & \textbf{77.3$\pm$9.1}     & 76.9$\pm$9.8              \\
    SMG Rt                    & 63.2$\pm$16.8            & 69.5$\pm$11.7             & 75.2$\pm$9.4              & \textbf{76.1$\pm$9.0}     \\
    Temporal Lobe Lt          & 66.7$\pm$3.6             & 80.9$\pm$3.7              & \textbf{82.6$\pm$6.4}     & 81.0$\pm$5.2              \\
    Temporal Lobe Rt          & 65.1$\pm$5.1             & 73.6$\pm$17.4             & \textbf{82.4$\pm$5.7}     & 80.5$\pm$4.0              \\
    Thyroid Lt                & 64.9$\pm$18.9            & 76.7$\pm$7.7              & 81.2$\pm$6.1              & \textbf{81.6$\pm$5.0}     \\
    Thyroid Rt                & 64.4$\pm$17.7            & 77.0$\pm$6.0              & 80.5$\pm$10.5             & \textbf{82.2$\pm$5.1}     \\
    TMjoint Lt                & 79.2$\pm$6.5             & 77.2$\pm$6.5              & \textbf{79.3$\pm$12.8}    & 77.6$\pm$7.0              \\
    TMjoint Rt                & 76.5$\pm$8.8             & 75.2$\pm$9.3              & \textbf{77.4$\pm$9.6}     & 76.2$\pm$7.1              \\ \hline
    Average                   & 66.6                     & 67.6                      & 70.4                      & \textbf{75.1}                     
    \end{tabular}
    }
    \caption{Dice score comparison on the  \ac{hn} 42 \ac{oar} dataset (unit: \%): Lt is short for left and Rt is short for right. Const. is short for constrictor muscle, SMG is short for submandibular gland, and TMjoint is short for temporomandibular joint. The proposed SOARS achieved the best performance in 30 (in bold) out of 42 \acp{oar}.}
    \label{tab:dice}
    \end{table}

    \subsection*{Performance of S\&H OAR detection}
    \label{sec:detect}
    
    In \Tab~\ref{tab:detect}, we report the category-by-category detection accuracy of the regressed center points using the detection-by-segmentation network. Moreover, we binaries both the regressed and ground-truth heat maps by keeping the top 1000 largest intensity voxels, and report their \ac{hd}. Note, as cochlea is spatially enclosed by inner-ear, we use a single heat map, \ie ear, for both \acp{oar} detection. As shown in \Tab~\ref{tab:detect},  we achieve an average \ac{hd} reduction of 13.7 mm (from 18.9 mm to 6.2 mm) as compared to the detection using only \ac{RTCT} images. The \ac{hd} for all \acp{oar} are reduced, especially the lacrimal gland, optic chiasm, and pineal gland. These significant \ac{hd} reductions indicate that the anchor \acp{oar} serve as effective references to better detect the \ac{sh} \ac{oar} locations. 
    
    \begin{table}[ht!]
    \centering
    \scalebox{0.7}{
    \begin{tabular}{lllll}
    Organ            & \multicolumn{1}{c}{UNet} & \multicolumn{1}{c}{P-HNN} & \multicolumn{1}{c}{UaNet} & \multicolumn{1}{c}{SOARS} \\ \hline
    Basal Ganglia Lt & 10.0$\pm$2.8             & 9.8$\pm$3.2               & 10.5$\pm$4.0              & \textbf{9.3$\pm$3.2}      \\
    Basal Ganglia Rt & \textbf{9.3$\pm$3.8}     & 10.2$\pm$3.3              & 10.5$\pm$3.8              & 11.1$\pm$3.4              \\
    Brachial Lt      & \textbf{14.9$\pm$6.2}    & 15.1$\pm$9.6              & 14.2$\pm$11.7             & 17.3$\pm$10.9             \\
    Brachial Rt      & 17.9$\pm$8.2             & \textbf{11.4$\pm$5.0}     & 16.2$\pm$9.6              & 14.0$\pm$7.3              \\
    Brainstem        & 8.4$\pm$2.9              & 8.8$\pm$2.9               & 10.3$\pm$3.8              & \textbf{8.1$\pm$2.2}      \\
    Cerebellum       & 8.9$\pm$3.8              & 9.4$\pm$4.7               & 14.1$\pm$9.8              & \textbf{7.7$\pm$3.1}      \\
    Cochlea Lt       & 3.6$\pm$9.0              & 1.8$\pm$0.5               & 2.3$\pm$0.8               & \textbf{1.6$\pm$0.4}      \\
    Cochlea Rt       & 2.1$\pm$0.8              & 2.0$\pm$1.0               & 2.4$\pm$0.9               & \textbf{1.9$\pm$0.6}      \\
    Const. inf       & 5.7$\pm$2.6              & 8.5$\pm$3.9               & 7.5$\pm$4.9               & \textbf{5.4$\pm$2.4}      \\
    Const. mid       & \textbf{7.4$\pm$2.8}     & 8.7$\pm$3.1               & 14.7$\pm$10.1             & \textbf{7.4$\pm$3.3}               \\
    Const. sup       & 7.4$\pm$3.0              & 8.0$\pm$3.6               & 12.7$\pm$8.2              & \textbf{7.0$\pm$3.6}      \\
    Epiglottis       & \textbf{6.7$\pm$2.3}     & 6.9$\pm$3.6               & 9.9$\pm$8.5               & 6.9$\pm$2.5               \\
    Esophagus        & 25.1$\pm$26.4            & 21.9$\pm$13.7             & 24.0$\pm$15.0             & \textbf{21.1$\pm$15.8}    \\
    Eye Lt           & \textbf{2.8$\pm$0.8}     & 3.0$\pm$1.8               & 4.0$\pm$5.4               & 3.3$\pm$1.1               \\
    Eye Rt           & 3.1$\pm$0.9              & 3.4$\pm$0.9               & 3.1$\pm$0.7               & \textbf{3.0$\pm$1.0}      \\
    Hippocampus Lt   & \textbf{11.0$\pm$6.7}    & 16.9$\pm$8.6              & 15.9$\pm$8.9              & 12.2$\pm$7.7              \\
    Hippocampus Rt   & \textbf{10.7$\pm$6.1}    & 12.7$\pm$5.8              & 13.3$\pm$6.6              & 12.5$\pm$8.2              \\
    Hypothalamus     & 16.9$\pm$8.6             & 9.3$\pm$4.3               & 10.3$\pm$3.7              & \textbf{2.5$\pm$1.3}      \\
    Innerear Lt      & 12.7$\pm$5.8             & 11.9$\pm$33.7             & 4.0$\pm$1.4               & \textbf{2.6$\pm$0.7}      \\
    Innerear Rt      & 9.3$\pm$4.3              & 4.1$\pm$1.3               & 4.7$\pm$2.8               & \textbf{2.9$\pm$0.8}      \\
    Lacrimal Gland Lt & 4.3$\pm$1.0              & 4.3$\pm$1.3               & 4.6$\pm$1.6               & \textbf{2.9$\pm$1.1}      \\
    Lacrimal Gland Rt & 4.1$\pm$1.2              & 5.5$\pm$1.5               & 5.1$\pm$2.2               & \textbf{2.9$\pm$0.9}      \\
    Larynx core          & 12.4$\pm$7.3             & 10.4$\pm$7.3              & 9.2$\pm$7.2               & \textbf{9.0$\pm$7.1}      \\
    Mandible Lt      & 7.9$\pm$2.9              & 6.7$\pm$2.8               & 10.3$\pm$24.4             & \textbf{5.3$\pm$2.3}      \\
    Mandible Rt      & 7.0$\pm$2.6              & 5.6$\pm$2.3               & 12.2$\pm$15.8             & \textbf{5.5$\pm$1.6}      \\
    Optic Chiasm     & 8.0$\pm$3.9              & 8.4$\pm$5.3               & 11.4$\pm$7.8              & \textbf{5.3$\pm$4.2}      \\
    Optic Nerve Lt   & 4.2$\pm$3.6              & 4.6$\pm$3.5               & 5.2$\pm$3.1               & \textbf{3.4$\pm$1.9}      \\
    Optic Nerve Rt   & 4.1$\pm$2.3              & 3.9$\pm$1.7               & 4.9$\pm$4.2               & \textbf{3.3$\pm$1.4}      \\
    Oralcavity       & 16.4$\pm$5.0             & 18.4$\pm$5.0              & 7.6$\pm$10.3             & \textbf{13.8$\pm$6.2}     \\
    Parotid Lt       & 9.0$\pm$3.4              & 10.0$\pm$2.8              & 8.0$\pm$5.8              & \textbf{7.0$\pm$2.5}      \\
    Parotid Rt       & 8.9$\pm$7.8              & 8.3$\pm$2.0               & 9.7$\pm$4.2               & \textbf{6.8$\pm$1.6}      \\
    Pineal Gland     & 3.4$\pm$1.8              & 2.5$\pm$1.1               & 4.0$\pm$1.9               & \textbf{1.7$\pm$0.6}      \\
    Pituitary        & \textbf{3.9$\pm$1.4}     & 4.4$\pm$1.6               & 4.4$\pm$1.3               & 4.2$\pm$2.2               \\
    Spinalcord       & 34.9$\pm$13.9            & 10.2$\pm$18.1             & 17.3$\pm$27.2             & \textbf{5.7$\pm$2.2}      \\
    SMG Lt           & 7.3$\pm$4.0              & 18.6$\pm$30.3             & \textbf{6.1$\pm$5.4}      & 6.5$\pm$3.1               \\
    SMG Rt           & 7.3$\pm$4.0              & 11.1$\pm$8.3              & 7.0$\pm$4.9               & \textbf{6.1$\pm$2.3}      \\
    Temporal Lobe Lt & \textbf{14.3$\pm$21.4}   & 16.0$\pm$6.8              & 16.5$\pm$6.7              & 14.6$\pm$6.9              \\
    Temporal Lobe Rt & \textbf{12.8$\pm$3.6}    & 38.6$\pm$85.2             & 15.0$\pm$5.0              & 13.5$\pm$5.9              \\
    Thyroid Lt       & 9.0$\pm$2.9              & 6.9$\pm$3.2               & 7.4$\pm$4.8               & \textbf{5.1$\pm$2.5}      \\
    Thyroid Rt       & 8.7$\pm$10.4             & 7.9$\pm$3.3               & 7.1$\pm$4.0               & \textbf{5.5$\pm$2.3}      \\
    TMjoint Lt       & \textbf{3.5$\pm$1.2}     & 3.9$\pm$1.4               & 4.4$\pm$2.4               & 3.6$\pm$1.7               \\
    TMjoint Rt       & 3.6$\pm$1.7              & 4.6$\pm$1.1               & 4.3$\pm$2.9               & \textbf{3.5$\pm$1.3}      \\ \hline
    Anchor OARs      & 9.3                      & 9.4                       & 9.2                       & \textbf{7.0}             
    \end{tabular}
    }
    \caption{Average Hausdorff distance comparison on the \ac{hn} 42 \ac{oar} dataset (unit: mm): Lt is short for left and Rt is short for right. Const. is short for constrictor muscle, SMG is short for submandibular gland, and TMjoint is short for temporomandibular joint. The proposed SOARS achieved the best performance in 30 (in bold) out of 42 \acp{oar}.}
    \label{tab:hd}
    \end{table}
    
    % The $1^{st}$ row is the \ac{RTCT} image with \ac{oar} delineations of a radiation oncologist. The $2^{nd}$ row shows \ac{oar} segmentation using the UaNet \acp{oar}. The $3^{rd}$ demonstrates the impact of \ac{nas}, indicating the necessity of adapting network architectures for different \acp{oar}.
    
    % \ac{soars} performed particularly better on anatomies that are difficult to segment due to the complexity in spatial structures, e.g., branchial, hippocampus, hypothalamus, lacrimal gland, oral cavity, optic chiasm, and pineal gland. With the help of anchor \acp{oar} predictions, \ac{soars} has better guidelines in anatomies' spatial location and therefore leads to better segmentation performance (from 66.5\% to 73.7\%). \ac{soars} performed slightly worse than UaNet on temporal lobe and TM joint segmentation. Yet, the \ac{dsc} differences are relatively small.
    
    % The advantage of \ac{soars} over the state-of-the-art UaNet is even more obvious when evaluated in terms of the \ac{hd}.  Without the \ac{oar} stratification, UaNet is prone to generate false positives especially when the \ac{oar} is relative small, e.g., hypothalamus, optic chiasm, and pineal gland. This is because the segmentation of UaNet is conducted in a segmentation-by-detection pattern, where the accuracy of detection has a great impact on the segmentation performance. Compared to the UaNet, which using only \ac{RTCT} for detection, \ac{soars} shows better detection accuracy, which leads to less false positives. 

    \begin{table}[ht]
    \centering
    \scalebox{0.75}{
    \begin{tabular}{llllll}
    \multicolumn{1}{c}{} & \multicolumn{2}{c}{Dist (mm)}                                    &  & \multicolumn{2}{c}{HD (mm)}                                      \\ \cline{2-3} \cline{5-6} 
    \multicolumn{1}{c}{} & \multicolumn{1}{c}{CT Only} & \multicolumn{1}{c}{CT+Anchor} &  & \multicolumn{1}{c}{CT Only} & \multicolumn{1}{c}{CT+Anchor} \\ \hline
    Ear Lt               & \textbf{3.9$\pm$2.5}        & 3.9$\pm$2.6                   &  & 6.7$\pm$3.3                 & \textbf{5.7$\pm$2.1}          \\
    Ear Rt               & 1.9$\pm$1.4                 & \textbf{1.6$\pm$1.0}          &  & 4.4$\pm$1.8                 & \textbf{3.4$\pm$1.3}          \\
    Hypothalamus         & 2.6$\pm$1.7                 & \textbf{2.3$\pm$1.5}          &  & 4.0$\pm$2.0                 & \textbf{3.6$\pm$1.5}          \\
    Lacrimal Gland Lt     & 5.6$\pm$5.7                 & \textbf{4.6$\pm$3.1}          &  & 28.0$\pm$76.8               & \textbf{14.7$\pm$20.7}        \\
    Lacrimal Gland Rt     & 3.3$\pm$1.9                 & \textbf{3.0$\pm$1.7}          &  & 47.4$\pm$112.0              & \textbf{4.7$\pm$1.4}          \\
    Optic Chiasm         & 3.9$\pm$2.5                 & \textbf{3.4$\pm$1.9}          &  & 26.6$\pm$71.8               & \textbf{10.6$\pm$25.6}        \\
    Optic Nerve Lt       & \textbf{2.5$\pm$1.6}        & 2.6$\pm$1.5                   &  & 4.6$\pm$1.8                 & \textbf{4.5$\pm$1.2}          \\
    Optic Nerve Rt       & \textbf{3.0$\pm$1.2}        & 3.1$\pm$1.6                   &  & 21.9$\pm$61.0               & \textbf{4.9$\pm$1.6}          \\
    Pineal Gland         & 2.5$\pm$2.5                 & \textbf{1.8$\pm$0.7}          &  & 27.7$\pm$72.2               & \textbf{3.9$\pm$1.3}          \\ \hline
    Average              & 3.3                         & \textbf{2.9}                  &  & 18.9                        & \textbf{6.2}                 
    \end{tabular}
    }
    \caption{The detailed \ac{sh} detection results measuring the average distances between regressed and true center points, as well as the Hausdorff distances between the binarised regressed and binarised true heat maps. Lt is short for left and Rt is short for right. The best performance is highlighted in bold. }
    \label{tab:detect}
    \end{table}
    
\end{document}